\documentclass[conference,letterpaper,10pt]{ieeeconf}
\usepackage[left=1.91cm,right=1.91cm,top=1.91cm,bottom=1.91cm]{geometry}

\usepackage{graphicx}
\usepackage{amssymb}
\usepackage{amsmath}
\usepackage{cite}
\usepackage{comment}
\usepackage{color}
\usepackage{url}
\usepackage{alltt}
\usepackage{cleveref}
\usepackage{tikz}
\usepackage{pgfplots}
\usepackage{pgf-umlsd}
\usepackage{cprotect}
\usepackage{multirow}
\usepackage{algorithm}
\usepackage[noend]{algpseudocode}
\usepackage{bigfoot}
\usepackage{paralist}
\usepackage{tabularx}
\usepackage[whole]{bxcjkjatype}
\usepackage{ifthen}
\usepackage{threeparttable}
\usepackage[hang,small,bf]{caption}
\usepackage[subrefformat=parens]{subcaption}
\usepackage[super]{nth}
\usepackage{colortbl}
\usepackage{scalefnt}
\usepackage{threeparttable}

\pgfplotsset{compat=1.14}
\IEEEoverridecommandlockouts

\makeatletter
\newcommand{\figcaption}[1]{\def\@captype{figure}\caption{#1}}
\newcommand{\tblcaption}[1]{\def\@captype{table}\caption{#1}}
\makeatother

\title{\LARGE \bf
Two-fingered Hand with Gear-type Synchronization Mechanism with Magnet
for Improved Small and Offset Objects Grasping: F2 Hand
}

\author{Naoki Fukaya$^{1}$, Avinash Ummadisingu$^{1}$, Kuniyuki Takahashi$^{1}$, Guilherme Maeda$^{1}$ and Shin-ichi Maeda$^{1}$
	\thanks{$^{1}$1 Preferred Networks, Inc.
  (As of the submission of this paper, Guilherme Maeda is affiliated with Sony AI.)
		{\small \{fukaya, ummavi, takahashi, ichi\}@preferred.jp,guilherme.maeda@sony.com}}
}

\begin{document}

\onecolumn

© 2023 IEEE.  Personal use of this material is permitted.  Permission from IEEE must be obtained for all other uses, in any current or future media, including reprinting/republishing this material for advertising or promotional purposes, creating new collective works, for resale or redistribution to servers or lists, or reuse of any copyrighted component of this work in other works.

\twocolumn

\newpage

\maketitle
%%%%%%%%%%%%%%%%%%%%%%%%%%%%%%%%%%%%%%%%%%%%%%%%%%%%%%%%%%%%%%%%%%%%%%%%%%%%%%%%%%%%%%%%%%%%%%%%%
%%%%%%%%%%%%%%%%%%%%%%%%%%%%%%%%%%%%%%%%%%%%%%%%%%%%%%%%%%%%%%%%%%%%%%%%%%%%%%%%%%%%%%%%%%%%%%%%%
\begin{abstract}

A problem that plagues robotic grasping is the misalignment of the object and gripper due to difficulties in precise localization, actuation, etc.
Under-actuated robotic hands with compliant mechanisms are used to adapt and compensate for these inaccuracies. However, these mechanisms come at the cost of controllability and coordination. For instance, adaptive functions that let the fingers of a two-fingered gripper adapt independently may affect the coordination necessary for grasping small objects.
In this work, we develop a two-fingered robotic hand capable of grasping objects that are offset from the gripper's center, while still having the requisite coordination for grasping small objects via a novel gear-type synchronization mechanism with a magnet.
This gear synchronization mechanism allows the adaptive finger's tips to be aligned enabling it to grasp objects as small as toothpicks and washers. The magnetic component allows this coordination to automatically turn off when needed, allowing for the grasping of objects that are offset/misaligned from the gripper. This equips the hand with the capability of grasping light, fragile objects (strawberries, creampuffs, etc.) to heavy frying pan lids, all while maintaining their position and posture which is vital in numerous applications that require precise positioning or careful manipulation.
\footnote{An accompanying video is available at the following link:\\ \url{https://www.youtube.com/watch?v=RAO7Qb2ZGNs}}

\end{abstract}
%%%%%%%%%%%%%%%%%%%%%%%%%%%%%%%%%%%%%%%%%%%%%%%%%%%%%%%%%%%%%%%%%%%%%%%%%%%%%%%%%%%%%%%%%%%%%%%%%
%%%%%%%%%%%%%%%%%%%%%%%%%%%%%%%%%%%%%%%%%%%%%%%%%%%%%%%%%%%%%%%%%%%%%%%%%%%%%%%%%%%%%%%%%%%%%%%%%

\section{INTRODUCTION}
\label{sec:Introduction}

A vital part of the robotic grasping pipeline is the accurate localization of the object and gripper. However, this may be challenging due to numerous factors like object recognition errors, localization errors driven by sensor limitations, noise, partial information, etc., or difficulty in moving the gripper precisely due to the complexity of control, actuation noise, operator errors, etc. Although precise control may be achieved by an operator carefully performing the task, it comes with the additional cost of time and undue effort.

A popular way to address this issue is the use of under-actuated robot hands~\cite{kim2020_hand,geies_adaptive_hand2021, Krug_belt_type_2fing_girpper2014} or soft robot hands~\cite{kuppuswamySoftbubbleGrippersRobust2020, dang2021robotic} which provide adaptive/compliance mechanisms that are able to compensate for these errors. This is accomplished through the use of compliant mechanisms which allow the hand to adapt to the object's position and grasp it stably without requiring high precision position control. Some compliant mechanisms also distribute contact force over a wider area allowing for a potentially more stable grasp and reducing the damage done to fragile objects with easily damageable surfaces such as fruit.

\begin{figure}[t]
    \centering
    \includegraphics[width=0.9\columnwidth]{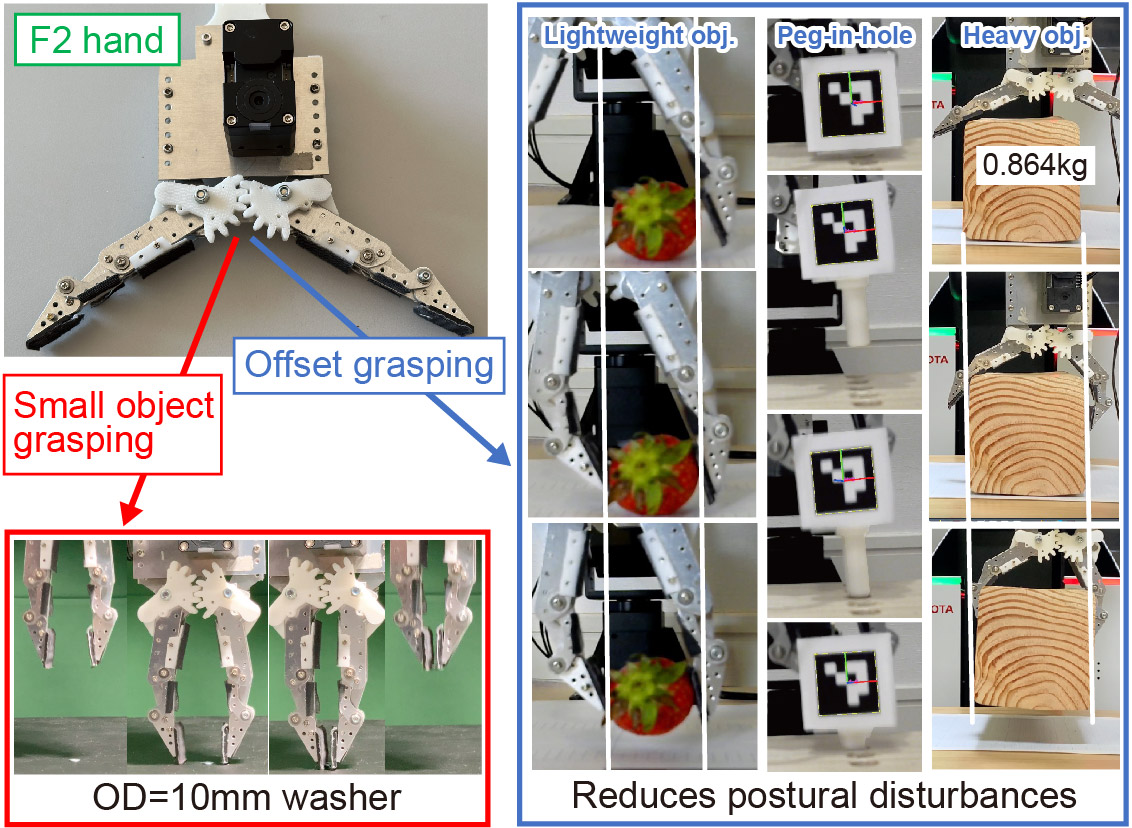}
    \caption{
    The F2 hand can grasp 1) small objects and 2) objects without changing their position and posture.
    Furthermore, this hand switches between these two functions automatically.
    }
    \label{fig:topfigure}
\end{figure}

Compliance mechanisms are typically designed so the fingers of the hand are able to adapt independently during the grasp to ensure a robust grasp is possible at the desired position. This is of particular consequence when trying to grasp objects that cannot be easily (re)positioned to accurately lie in the center of the fingers (due to constraints in movement, positioning, fragility, etc.). However, this independence comes with unique challenges in terms of coordination or synchronicity of fingertips during grasp closure and a lack of coordination between the fingers may lead to their misalignment when the hand is closed during grasping. This may cause a deterioration in the performance of grasping small objects that require a precise pinch grasp. 

On the other hand, if fingers are synchronized to close too strongly, objects that are offset from the hand's center end up getting dragged along the surface during grasping which may be undesirable if they are fragile. Additionally, objects that cannot be dragged or displaced in posture (e.g., peg-in-hole task or a heavy object for which the object's static frictional force is greater than the gripper's grasping force) cannot be grasped because the gripper cannot close.

Therefore, in this work, we aim to design a robotic hand that can achieve robust grasping and be potentially applied to a wide range of tasks in the face of object localization and end-effector positioning errors that satisfy the following criteria:
1) To be able to grasp small objects, which require accurate alignment of the fingertips, and 2) adaptive grasping (hereafter referred to as \emph{offset grasping}), in which object to be grasped, including fragile and heavy objects, is offset from the hand's center, without changing its position and posture.

To realize these objectives, we propose the following novel synchronization mechanisms using two fingers themselves using adaptive mechanisms that are inspired by human fingers (Fig.~\ref{fig:topfigure}).

\begin{itemize}
\item A gear mechanism with a synchronization mechanism in which the adaptive fingers' tips stay aligned and fingers are synchronized during grasp closure.
\item A mechanism driven by magnets that automatically switches between two grasping modes enabling grasping of small objects via a synchronized pinch grasp and offset grasping.  
\end{itemize}

%%%%%%%%%%%%%%%%%%%%%%%%%%%%%%%%%%%%%%%%%%%%%%%%%%%%%%%%%%%%%%%%%%%%%%%%%%%%%%%%%%%%%%%%%%%%%%%%%
%%%%%%%%%%%%%%%%%%%%%%%%%%%%%%%%%%%%%%%%%%%%%%%%%%%%%%%%%%%%%%%%%%%%%%%%%%%%%%%%%%%%%%%%%%%%%%%%%
\section{RELATED WORK}
\label{sec: related_work}

%%%%%%%%%%%%%%%%%%%%%%%%%%%%%%%%%%%%%%%%%%%%%%%%%%%%%%%%%%%%%%%%%%%%%%%%%%%%%%%%%%%%%%%%%%%%%%%%%
Many robot hands with adaptive functions, such as under-actuated robot hands, have been developed in the past due to their versatility.
Soft robotic hands are used for grasping fragile objects such as foodstuff because the fingers or the whole hand are made with soft materials and typically grasp in a way to distribute pressure across as large an area as possible~\cite{dimeas2013towards,giannaccini_single_softgripper_2014, endo2020robotic,wang2021robotic}.
Typically, when fingers and palms are made with such soft materials, the positional precision of the fingertips is reduced, making it challenging to perform fine-grained grasping movements such as pinching when trying to grasp small or thin objects.
Since such manipulation is one of the most important functions for a gripper, several prior works explored the use of adaptive hands to grasp~\cite{kota_watanabe_belt_2fin_gripper_2020,KoTendonDrivenPFN,odhner_adaptive_precision_grasp_2012,laliberte_underacutation_hand_2007}.
\cite{teeple2020_hand} made multiple chambers in the fingers of a soft gripper to enable both pinching and wrapping like a power grasp.
Many under-actuated robotic hands made of common materials such as plastic and metal are also made to be able to perform a pinching grasp, and \cite{GaoAdaptiveFinger2016} has achieved both wrapping and pinching grasps by applying an adaptive function to parallel gripper fingers. 
\cite{watanabe2021_hand} studied how to grasp small objects (e.g., $2\mathrm{\,mm}$ thick cushions and $0.7\mathrm{\,mm}$ thick rubber sheets) by using a finger with an adaptive function.

To grasp a small object between opposable fingers, as is done in this study, the fingertip positions of both fingers must match.
In such a grasping, the fingers usually move toward the center of the hand.
Taking advantage of the wrap-around grasping, the ability to grasp even when the relative position of the object and the hand is significantly misaligned has also been studied.
\cite{dollar_sdm_hand_2010} has developed a structure that allows the wrist to flex to accommodate grasping even when the object is significantly misaligned. However, this structure causes the position of the object to shift after grasping, which may be undesirable. 
\cite{DefGear_qiujie_gripper_2021} has also developed a structure that uses differential gears to interlock the left and right fingers to grasp the object, even if the position of the grasped object is offset. However, this structure does not keep the position and posture of the object after grasping the same, and it may shift after grasping.
Moreover, this gripper cannot grasp small objects due to a lack of fingertip alignment.
Such a situation where the object position changes after grasping may adversely affect the robot's ability to continue working with the object (such as placing) and would require a re-estimation of its now-changed pose. If the adaptive fingers' tips stay aligned and fingers are synchronized during grasp closure, or the change in object pose after grasping can be suppressed, the need for such pose compensation or re-estimation would be reduced.

%%%%%%%%%%%%%%%%%%%%%%%%%%%%%%%%%%%%%%%%%%%%%%%%%%%%%%%%%%%%%%%%%%%%%%%%%%%%%%%%%%%%%%%%%%%%%%%%%
%%%%%%%%%%%%%%%%%%%%%%%%%%%%%%%%%%%%%%%%%%%%%%%%%%%%%%%%%%%%%%%%%%%%%%%%%%%%%%%%%%%%%%%%%%%%%%%%%
\section{HAND DESIGN}

To realize both small objects and offset grasping capabilities, we developed a robot hand that combines the adaptive grasping function with an adaptive mechanism and the small object grasping function using a gear-type synchronization mechanism with a magnet.

%%%%%%%%%%%%%%%%%%%%%%%%%%%%%%%%%%%%%%%%%%%%%%%%%%%%%%%%%%%%%%%%%%%%%%%%%%%%%%%%
\subsection{Adaptive Fingers}

\begin{figure}[t]
	\centering
	\includegraphics[width=0.75\columnwidth]{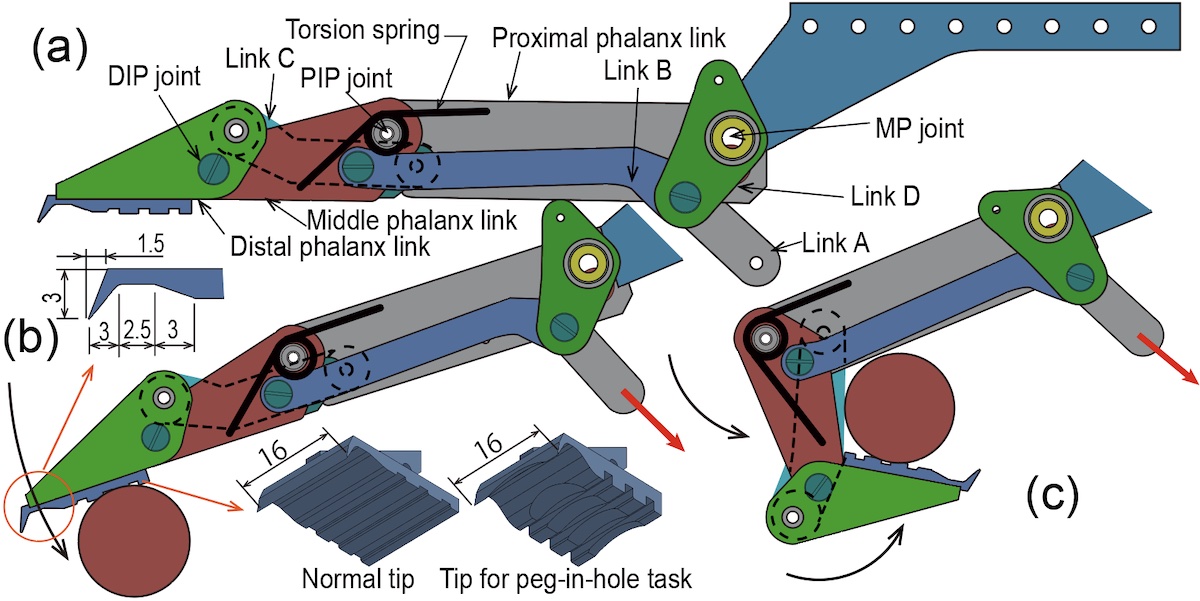}
	\caption{
    Structure of a finger with an adaptive mechanism.
    }
	\label{fig:fingerworks}
\end{figure}

To make the proposed robot hand, we decided to use a finger structure with an adaptive function that enables intuitive grasping of various objects without the use of any sensors that were developed and used by the authors in prior work~\cite{Fukaya_F3_hand,gmaeda_f1hand_2022, fukaya2000_hand_B}. An overview of this finger mechanism is shown in Fig.~\ref{fig:fingerworks}. 
This finger was developed to realize typical human grasping motions with simple operability. Depending on the state of contact between the object and the finger, the finger can automatically use either a pinch grasp, in which the finger remains extended, or an adaptive grasp, in which the finger makes contact to wrap the object. These grasping modes were inspired by typical motions made by humans fingers intuitively when faced with similar tasks. When the distal phalangeal link contacts the object, the rotating finger transmits the grasping force while keeping extension (Fig.~\ref{fig:fingerworks}~(b)). When the proximal phalangeal link reaches the object, link D rotates independently, activating the middle phalangeal link via link B, which pulls link C (connected to the proximal phalangeal link) to pull the distal phalangeal link (Fig.~\ref{fig:fingerworks}~(c)). As a result, the movable finger flexes to wrap the object. The grasping motion stops automatically when all possible contact areas are contacted. At this point, the contact area is maximized wherever possible. Since the contact force is distributed over the entire contact area, the object can be grasped stably even with a small motor torque.
The nails were attached to the fingertips, as shown in Fig.~\ref{fig:fingerworks}, and a non-slip sheet TB631 (3M) was attached to the finger surface.
Finger dimensions were determined based on the author's middle finger. The width was $16\mathrm{\,mm}$, and can be changed as desired.

%%%%%%%%%%%%%%%%%%%%%%%%%%%%%%%%%%%%%%%%%%%%%%%%%%%%%%%%%%%%%%%%%%%%%%%%%%%%%%%%
\subsection{Structure of F2 hand}
\label{sec:f2design}

%%%%%%%%%%%%%%%%%%%%%%%%%%%%%%%%%%%%%%%%%%%%%%%%%%%%%%%%%%%%%%%%%%%%%%%%%%%%%%%%%%%%%%%%%%%%%%%%%

\begin{figure}[t]
	\centering
	\includegraphics[width=0.93\columnwidth]{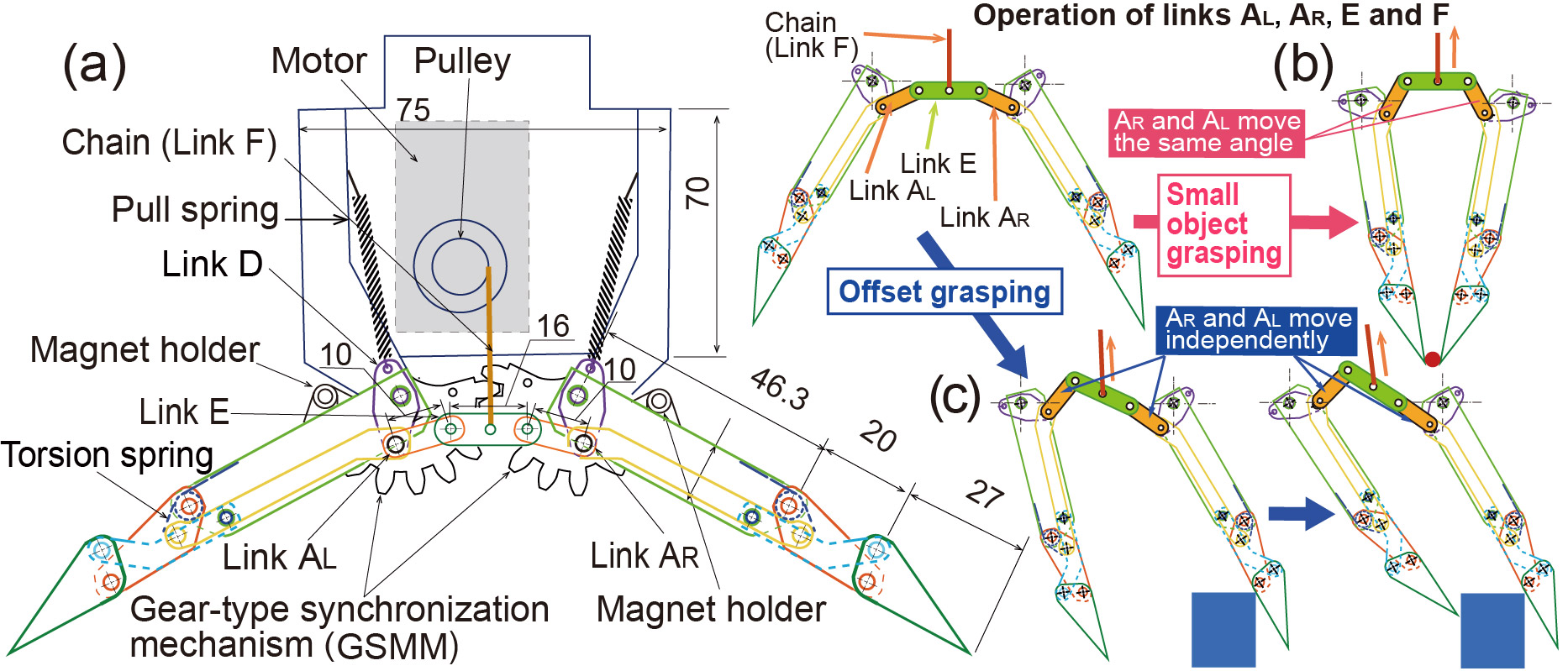}
	\caption{
	   Structure of the F2 hand. Fingers with adaptive mechanisms are joined via links $\rm{A_R}$, $\rm{A_L}$, E, and F.
        A gear synchronization mechanism is coaxially arranged on the fingers.
    }
	\label{fig:f2overview}
\end{figure}

We note that the authors have developed various hands that incorporate the adaptive finger used in this work, such as F1 hand~\cite{gmaeda_f1hand_2022} and F3 hand~\cite{Fukaya_F3_hand}.
The F1 hand~\cite{gmaeda_f1hand_2022 } has only one adaptive finger, making it challenging to perform offset grasping.
In the F3 hand~\cite{Fukaya_F3_hand}, the motor is mounted on the finger opposing the adaptive finger, so each finger must be operated individually.
Therefore, it is challenging to perform offset grasping, and the operation is more complex and not very intuitive.
These challenges motivated some design choices in developing the F2 hand~(Fig.~\ref{fig:f2overview}).

F2 hand consists of two adaptive fingers, link mechanisms that connect the fingers and distributes traction to each finger, a gear-type synchronization mechanism that moves the two fingers in a coordinated way, and a motor (Dynamixel XM430-350R) that pulls the link to bend the fingers.
The finger is pulled back to the initial position by a coil spring (spring constant $K=0.114\mathrm{\,N/mm}$) attached near the Metacarpophalangeal joint (MP joint). This spring was selected to have the minimum tension necessary for the finger to return to its initial position since a strong spring tension would adversely affect its offset grasping ability.
A torsion spring (spring constant $K=0.13\mathrm{\,N \cdot mm/deg}$) is attached to the Proximal Interphalangeal joint (PIP joint). This spring is used to maintain extension when the finger is flexed. For this reason, one with a slightly higher reaction force than a coil spring is chosen.
Materials of the main parts of F2 hand are A2017 with a thickness of $1.5\mathrm{\,mm}$, $2\mathrm{\,mm}$, and ABS. Weigh is $210\mathrm{\,g}$, including the motor.

%%%%%%%%%%%%%%%%%%%%%%%%%%%%%%%%%%%%%%%%%%%%%%%%%%%%%%%%%%%%%%%%%%%%%%%%%%%%%%%%
\subsection{Offset Grasping Functions}
\label{sec:ofset_grasping}

\begin{figure}[t]
	\centering
	\includegraphics[width=0.75\columnwidth]{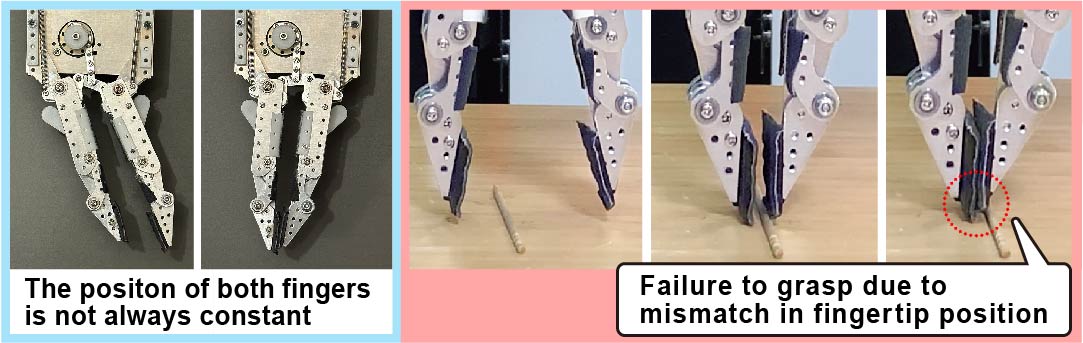}
	\caption{
        A situation in which the left and right fingers are not actuated in a coordinated manner.
    }
	\label{fig:badmotion}
\end{figure}

The driving force from the motor is transmitted to the link $\rm{A_R}$ \& $\rm{A_L}$ of each finger via link E and chain (link F). 
These links enable two behaviors of the gripper depending on the contact situation between the finger and the object.

Fig.~\ref{fig:f2overview} (right) shows its movement when only the fingers and link $\rm{A_R}$ \& $\rm{A_L}$, E and F are taken out and the chain (link F) is pulled by a motor.
If nothing is touching the object when the link F is pulled, the fingers on both sides rotate equally and the fingertips coincide at the center of the hand (Fig.~\ref{fig:f2overview}~(b)). This action is useful for grasping small objects.

In Fig.~\ref{fig:f2overview}~(c), even though the right fingertip contacts an object and link $\rm{A_R}$ is stationary, link E moves independently and pulls link $\rm{A_L}$.
Then, when the left fingertip contacts the object, link $\rm{A_L}$ comes to rest.
If the relative position of the object and fingers does not change, the link remains at rest, and it can be lifted the object in that position.

However, in this structure in which each adaptive finger or link $\rm{A_R}$, $\rm{A_L}$, E, and F moves independently, there are differences in the movement of each finger due to friction in the rotating parts, the reaction force of the spring, and the effect of gravity. For example, the fingertips of the left finger reach the center of the hand before the fingertips of the right finger sometimes. This misalignment of fingertips is disastrous when grasping small or thin objects such as toothpicks or washers, as shown in Fig.~\ref{fig:badmotion}. 
Therefore, a mechanism is incorporated to synchronize the movement of both adaptive fingers using gears.

%%%%%%%%%%%%%%%%%%%%%%%%%%%%%%%%%%%%%%%%%%%%%%%%%%%%%%%%%%%%%%%%%%%%%%%%%%%%%%%%
\subsection{Gear-type Synchronization Mechanism with Magnets}
\label{sec:gear_type_sync_mechanism}

\begin{figure}[t]
	\centering
	\includegraphics[width=0.99\columnwidth]{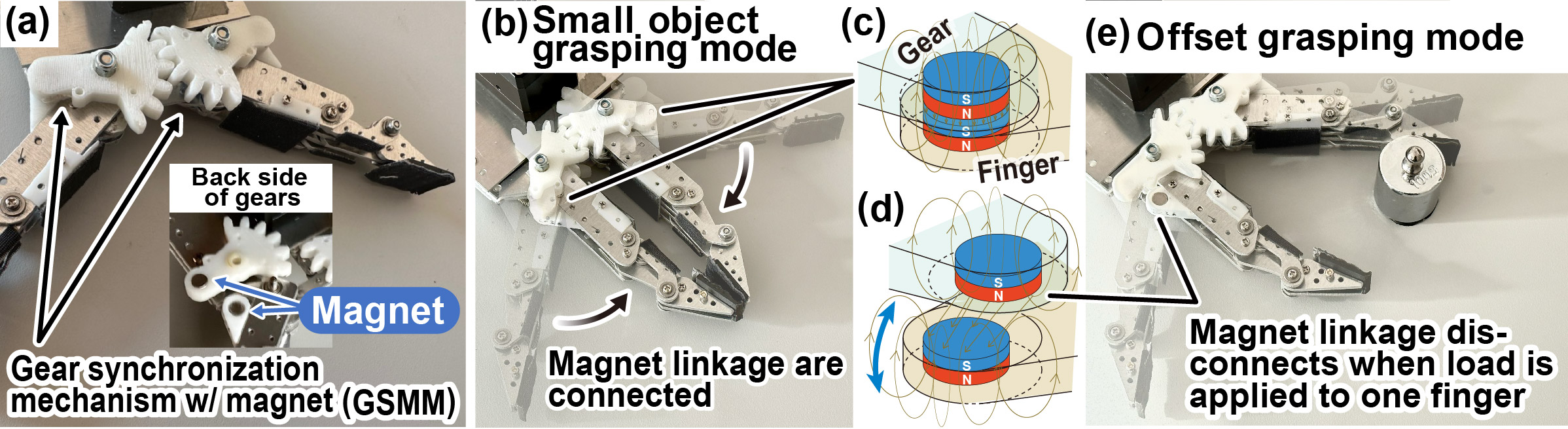}
	\caption{
             Gear-type synchronization mechanism with magnets (GSMM).
                }
	\label{fig:magnet_positon}
\end{figure}

%%%%%%%%%%%%%%%%%%%%%%%%%%%%%%%%%%%%%%%%%%%%%%%%%%%%%%%%%%%%%%%%%%%%%%%%%%%%%%%%

If the motion of the two fingers is simply synchronized by the mechanism, this can have a negative effect on offset grasping.
For instance, when only one finger is in contact with an object, the presence of this synchronized mechanism causes the two fingers to continue to flex in the same manner. This causes the object to move so that it is dragged along the surface toward the center of the hand until the other finger makes contact. This may cause further complications when the object is easily damaged or extremely rough and heavy. 
Therefore, we have developed a structure in which a gear is placed coaxially on the MP joint of the fingers, and this gear is connected to the fingers by a magnet so that the synchronization state of the two fingers can be changed as needed. We call this mechanism a gear-type synchronization mechanism with a magnet (GSMM) (Fig.~\ref{fig:magnet_positon}~(a)).

Two disc-shaped neodymium magnets of $6\mathrm{\,mm}$ in diameter and $3\mathrm{\,mm}$ thick were used in combination, whose magnetic flux density is $280\mathrm{\,mT}$ each.
The magnetic force attracts the two magnets so that their central axes coincide. 
This allows the left and right fingers to move synchronously until they close in the center, as long as no load is applied  (Fig.~\ref{fig:magnet_positon}~(b),~(c)).
When a load is applied, the magnets exhibit springiness until completely separated. Therefore, they return to their original position after the load is removed. When the motor is reversed, and the fingers return to their original position, the separated magnets are automatically joined together by magnetic force (Fig.~\ref{fig:magnet_positon}~(c),~(d)).
Since the fingers are hardly loaded during the closing motion, GSMM can be sufficiently interlocked by magnets. On the other hand, if a load is applied during the closing motion, such as when the fingers strike an object, the magnet leaves, and an offset grasp is performed (Fig.~\ref{fig:magnet_positon}~(d),~(e)). Since these actions are automatic and passive, the operator does not need to be aware of the switch in behavior.
Once disconnected, GSMM automatically returns to the magnet-locked state by magnetic force, returning it to the initial position by spreading the fingers. 
GSMM also has the effect of linking the contact force of the left and right fingers to the other finger.
The force required to release the lock can be adjusted by adjusting the magnetic force of the magnet.
This GSMM does not require additional actuators or controls and can be easily applied to hands with other passive structures.
This GSMM is one of the contributions of our research.

%%%%%%%%%%%%%%%%%%%%%%%%%%%%%%%%%%%%%%%%%%%%%%%%%%%%%%%%%%%%%%%%%%%%%%%%%%%%%%%%%%%%%%%%%%%%%%%%%
%%%%%%%%%%%%%%%%%%%%%%%%%%%%%%%%%%%%%%%%%%%%%%%%%%%%%%%%%%%%%%%%%%%%%%%%%%%%%%%%%%%%%%%%%%%%%%%%%
\section{EXPERIMENTAL SETUP}
\label{sec: experimental_conditions}

%%%%%%%%%%%%%%%%%%%%%
    As discussed in \cref{sec:Introduction}, the objective set in designing the F2 hand is to create a highly versatile robotic hand that can grasp objects in the presence of localization and positional errors. Additionally, it should be able to grasp small objects and perform grasps from an offset position wherever needed.
    To accurately validate the functionality of our hand and the GSSM mechanisms we aim to verify the following:
    P1) If the GSSM enables the precision grasping of small objects.
    P2) If the GSSM switch between the modes (offset and small object grasping) as and when required.
    P3) If the GSSM allows offset grasping without significantly changing the object's position on the surface or its angle.

%%%%%%%%%%%%%%%%%%%%%%%%%%%%%%%%%%%%%%%%%%%%%%%%%%%%%%%%%%%%%%%%%%%%%%%%%%%%%%
    \subsection{Modes of Operation}
    \label{sec: confirm_finger_syc_motoi}

%%%%%%%%%%%%%%%%%%%%%%%%%%%%%%%%%%%%%%%%%%%%%%%%%%%%%%%%%%%%%%%%%%%%%%%%%%%%%%

\begin{figure}[t]
	\centering
	\includegraphics[width=0.75\columnwidth]{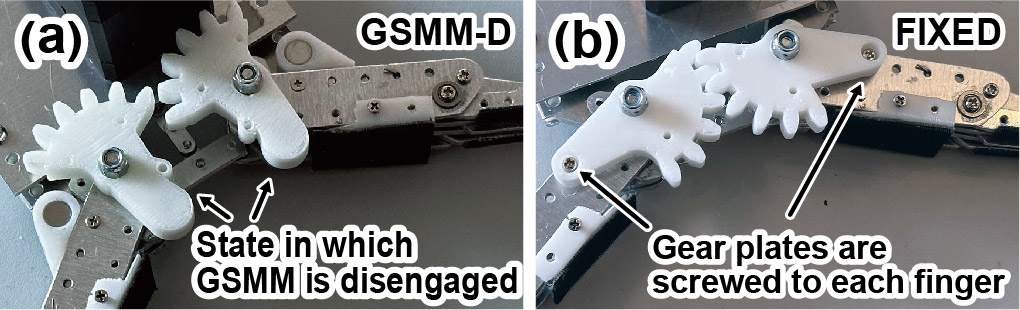}
	\caption{
            The state in which the GSMM-D and the FIXED.
    }
	\label{fig: grasping_motion_type}
\end{figure}

    In our experiments, we aim to evaluate the effects of the proposed GSSM and its components by evaluating the GSSM and two variants:
    \begin{itemize}
        \item[a)] GSMM: The GSSM is engaged and fingers are magnetically connected with the ability to automatically disengage if needed（proposed method)~(Fig.~\ref{fig:magnet_positon}).
        \item[b)] GSMM-Disengaged (GSMM-D): The GSSM is present but disengaged and the fingers are not synchronized~(Fig.~\ref{fig: grasping_motion_type}~(a)). It is equivalent to a system without GSMM and no finger synchronization linkages.
        \item[c)] GSSM-FIXED (FIXED): The gear-type synchronization mechanism is fixed to the finger with a screw so that the rotation of both MP joints is always synchronized ~(Fig.~\ref{fig: grasping_motion_type}~(b)). 
    \end{itemize}
    
%%%%%%%%%%%%%%%%%%%%%%%%%%%%%%%%%%%%%%%%%%%%%%%%%%%%%%%%%%%%%%%%%%%%%%%%%%%%%%%%
    \subsection{Experiment Environment}
    
\begin{figure}[t]
	\centering
	\includegraphics[width=0.95\columnwidth]{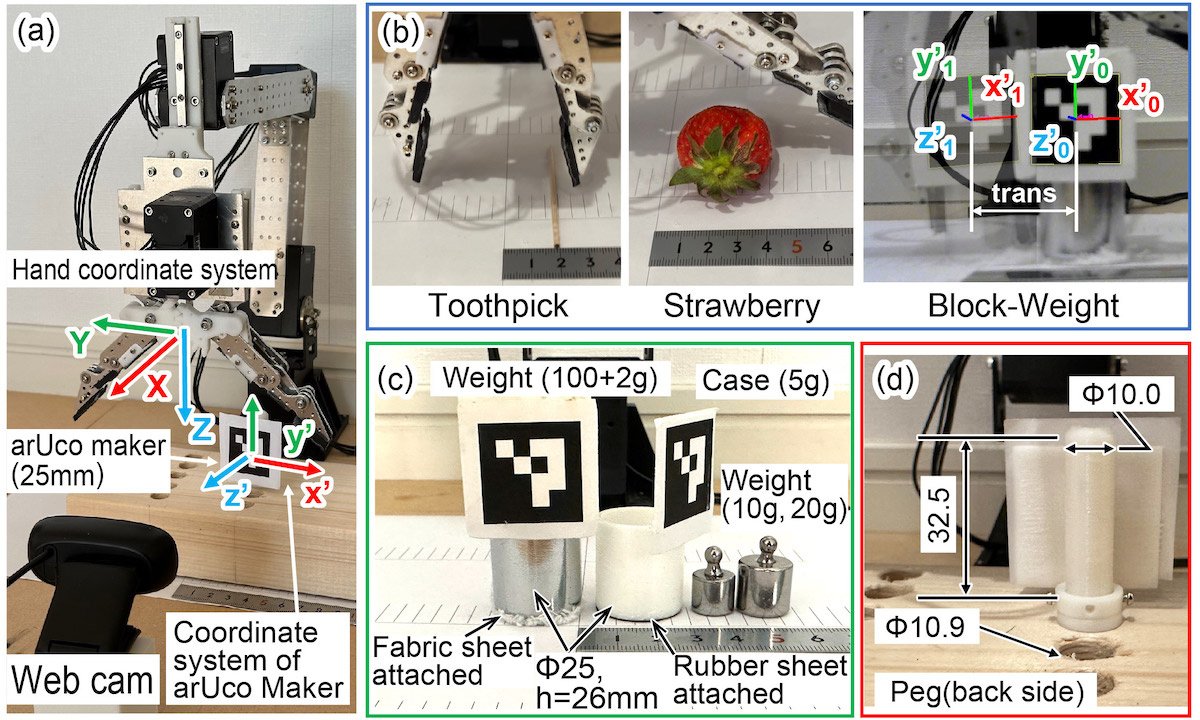}
	\caption{
        Equipment used to check basic grasping functions. 
        a) Mini-robot arm and respective coordinate system.
        b) Example of placement of the object to be measured.
        c) Weights used for the measurement.
        d) Basic shape of the Peg used in the measurement.
    }
	\label{fig:basic_arm}
\end{figure}

    The robot hand is equipped to a 4-axis mini-arm (Dynamixel XM430-W350-R$\times$2, XM540-W270-R$\times$2). During the experiment, it was allowed to move only along the Z-axis (in hand coordinates) up to $50\mathrm{\,mm}$ in a linear slider.
    Subsequent sections describe the different tasks and environments we evaluate the F2 hand in to examine if it achieves the expected performance and sufficiently verifies the points raised. The various objects that were grasped in our experiments, the configuration of the mini-robotic arm, and the arrangement of the objects are visualized in Fig.~\ref{fig:basic_arm}.

%%%%%%%%%%%%%%%%%%%%%%%%%%%%%%%%%%%%%%%%%%%%%%%%%%%%%%%%%%%%%%%%%%%%%%%%%%%%%%%%
    \subsection{Grasping of a Small Object}
    \label{sec: confirmation_small_item}
    
%%%%%%%%%%%%%%%%%%%%%%%%%%%%%%%%%%%%%%%%%%%%%%%%%%%%%%%%%%%%%%%%%%%%%%%%%%%%%%%%
    In order to verify P1 and confirm that the gear linkage mechanism improves performance when grasping small objects, we perform experiments on a toothpick placed on white paper~(Fig.~\ref{fig:basic_arm}~(b)). After ten grasp attempts, the toothpick is moved horizontally by $5\mathrm{\,mm}$, and another grasping experiment is performed. Grasping is performed using two modes- GSMM and GSMM-D (GSMM with magnetic synchronization disengaged).
    The same experiment is repeated with a parallel gripper for comparison. The parallel gripper is modeled after the one used in the study of a food-grasping robot~\cite{Takahashi_food_manipuration_2022, ummadisingu2022cluttered}. For a fair comparison, the fingertips were replaced with the same nail-like structure as those of the F2 hand.
    The Z-axis target of the hand during the grasp is set at a suitable position for grasping (a height slightly touching the floor for the parallel gripper and a height of about $-3\mathrm{\,mm}$ for the F2 hand, where the joint adaptive mechanism acts). This is based on the assumption that the fingertips contact the floor or an object by means of a wrist torque sensor, etc., and stop at a position suitable for grasping.

%%%%%%%%%%%%%%%%%%%%%%%%%%%%%%%%%%%%%%%%%%%%%%%%%%%%%%%%%%%%%%%%%%%%%%%%%%%%%%%%
    \subsection{Offset Grasping: Position Changes}
    \label{sec: confirmation_offset_grasping}
   
%%%%%%%%%%%%%%%%%%%%%%%%%%%%%%%%%%%%%%%%%%%%%%%%%%%%%%%%%%%%%%%%%%%%%%%%%%%%%%%%
    To verify P3 and the arm's ability to perform offset-grasping, we evaluate the outcomes of grasps executed on objects placed at increasing distances away from the central axis of the hand. 
    The object is placed on white paper with its rightmost edge being $20\mathrm{\,mm}$ from the central axis of the hand and moved in $5\mathrm{\,mm}$ increments in the X-axis direction until it reaches $60\mathrm{\,mm}$ (Fig.~\ref{fig:basic_arm}~(a) \& (b)). 
    Based on our past experience with this grasping setup, we assume a reasonable position error of 2 to $3\mathrm{\,mm}$, accounting for object detection and localization error by the RGB sensor during the grasp attempt.
    We verify the ability to perform offset grasping and measure the amount of lateral displacement of the weighted block when at least offset by a value greater than the above to simulate a more realistic grasp scenario.

%%%%%%%%%%%%%%%%%%%%%%%%%%%%%%%%%%%%%%%%%%%%%%%%%%%%%%%%%%%%%%%%%%%%%%%%%%%%%%%%
    \subsubsection{Offset grasping of block-weights}
    \label{sec: confirmation_offset_weight}

%%%%%%%%%%%%%%%%%%%%%%%%%%%%%%%%%%%%%%%%%%%%%%%%%%%%%%%%%%%%%%%%%%%%%%%%%%%%%%%%

    To quantitatively evaluate the hand's ability to perform offset grasps, we use weighted blocks with $25\mathrm{\,mm}$ arUco markers placed on them as a means to accurately measure the lateral displacement of the block during offset grasping. 
    We use two kinds of weighted blocks placed on paper for this experiment- one with a fabric attached to the base to make it somewhat slippery, and a case covered with a rubber sheet to achieve friction similar to that of a strawberry, to match another object of interest (Fig.~\ref{fig:basic_arm}~(b) \& (c)).
    
    The coefficient of friction of the fabric-applied block-weight is 0.4, and the rubber-applied case is 0.9. The weights selected were $10\mathrm{\,g}$ and $20\mathrm\,g$ , which is similar in weight to a strawberry, and they were placed in a hand-made case ($5\mathrm{\,g}$) similar in shape to that of the $100\mathrm{\,g}$ weight so that the contact area would be constant. 

%%%%%%%%%%%%%%%%%%%%%%%%%%%%%%%%%%%%%%%%%%%%%%%%%%%%%%%%%%%%%%%%%%%%%%%%%%%%%%%%
    \subsubsection{Offset grasping of fragile objects}
    \label{sec: confirmation_offset_strawberry}

%%%%%%%%%%%%%%%%%%%%%%%%%%%%%%%%%%%%%%%%%%%%%%%%%%%%%%%%%%%%%%%%%%%%%%%%%%%%%%%%

    We conduct additional grasping experiments on strawberries- a food susceptible to damage when dragged along the surface of the grasp to represent a difficult real-world task that motivates offset grasping.
    To verify the effect of the gear linkage mechanism, the same strawberries were grasped by GSMM and FIXED, and the differences in behavior and the amount of damage done to the strawberries were observed. The weight of the strawberry used was 17.4 g.

In order to further confirm the effect of offset grasping, experiments are also conducted on more foodstuffs including tomatoes, shiitake mushrooms, and cream puffs. A parallel gripper is also used for comparison.

%%%%%%%%%%%%%%%%%%%%%%%%%%%%%%%%%%%%%%%%%%%%%%%%%%%%%%%%%%%%%%%%%%%%%%%%%%%%%%%%

    \subsection{Offset Peg-in-hole: Position and Posture Changes}
    \label{sec: confirmation_peg_in_hoke}
%%%%%%%%%%%%%%%%%%%%%%%%%%%%%%%%%%%%%%%%%%%%%%%%%%%%%%%%%%%%%%%%%%%%%%%%%%%%%%%%

    To evaluate the postural displacement of the object during offset grasping, we attempt a variation of the classical peg-in-hole task of removing an inserted peg from its hole and reinserting it, from a position where the gripper is offset from the peg. We believe this function is vital in situations like removing vials of liquid from shelves without tilting them enough to spill under reachability constraints. 
    An ABS pin $10.0\mathrm{\,mm}$ in diameter and $65\mathrm{\,mm}$ long is inserted $25\mathrm{\,mm}$ into a hole $10.9\mathrm{\,mm}$ in diameter to check if it could be reinserted after being pulled out (Fig.~\ref{fig:basic_arm}~(d)).
    The peg is pulled up $40\mathrm{\,mm}$ in the Z-axis direction manually, and the peg is lowered again to see if it can be reinserted.
    The amount of movement in the X-direction from the initial position and the amount of rotation around the yaw axis was measured by the arUco marker attached to the pin and pulled out state.
    In this experiment, depressions are made on the tip surface to eliminate minor effects of horizontal displacement (Fig.~\ref{fig:fingerworks}).

%%%%%%%%%%%%%%%%%%%%%%%%%%%%%%%%%%%%%%%%%%%%%%%%%%%%%%%%%%%%%%%%%%%%%%%%%%%%%%%%
    \subsection{Autonomous Grasping}
    \label{sec: autonomous_grasping_conditions}

%%%%%%%%%%%%%%%%%%%%%%%%%%%%%%%%%%%%%%%%%%%%%%%%%%%%%%%%%%%%%%%%%%%%%%%%%%%%%%%%
    
    The motivating use of the proposed system is the ability to grasp a wide range of objects under localization and pose uncertainties of the object and gripper. To verify this, we construct an intentionally simplistic autonomous grasping pipeline consisting of a Sawyer 7-DoF robotic arm performing grasping from RGB images in an open-loop bin-picking style setup. 
    An IDS uEye RGB camera overlooks the workspace of the robot arm and acquires images of the grasped object.
    The workspace consists of a black styrene board ($485\mathrm{\,mm} \times 570\mathrm{\,mm}$).
    The Sawyer, the F2 hand's motor, and the RGB camera are connected to a PC running Ubuntu 16.04 and ROS Kinetic.
    At runtime, during initialization, the image of the styrene board workspace is imaged to create a reference image.
    Next, when the grasping object is placed on the styrene board, we use simple background subtraction methods to acquire a mask of the object.
    Using the foreground mask of the object, we estimate the grasp position as the centroid of the mask, and its Z-axis is fixed to the height at which the fingertip bends slightly when it contacts the surface of the workspace ($-3\mathrm{\,mm}$).

%%%%%%%%%%%%%%%%%%%%%%%%%%%%%%%%%%%%%%%%%%%%%%%%%%%%%%%%%%%%%%%%%%%%%%%%%%%%%%%%
    \subsection{Teleoperation using the HSR}
    \label{sec: mamual_manipulation_conditions}

    %%%%%%%%%%%%%%%%%%%%%%%%%%%%%%%%%%%%%%%%%%%%%%%%%%%%%%%%%%%%%%%%%%%%%%%%%%%%%%%%

    To further verify the performance of the arm in the presence of imprecise positioning, we perform teleoperation using a mobile manipulator (Toyota Human Support Robot (HSR)~\cite{yamamoto2019development}) controlled using a 3D mouse (Connexion SpaceMouse).
    The HSR, the F2 hand's motor, and the 3D mouse are connected to a PC running Ubuntu 18.04 and ROS melodic and run through ROS.
    The 3D mouse input is synchronized with three directions of the tool coordinate system in addition to the rotation θ around the Z-axis shown in Fig.~\ref{fig:wood_block}~(a).
    This allows the HSR's wrist joint to move in the same direction as the 3D mouse input.
    The speed of the end-effector movement is changed in proportion to the amount of input from the 3D mouse.

%%%%%%%%%%%%%%%%%%%%%%%%%%%%%%%%%%%%%%%%%%%%%%%%%%%%%%%%%%%%%%%%%%%%%%%%%%%%%%%%%%%%%%%%%%%%%%%%%
\section{EXPERIMENTAL RESULT}
\label{sec: experimental_result}
%%%%%%%%%%%%%%%%%%%%%%%%%%%%%%%%%%%%%%%%%%%%%%%%%%%%%%%%%%%%%%%%%%%%%%%%%%%%%%%%

%%%%%%%%%%%%%%%%%%%%%%%%%%%%%%%%%%%%%%%%%%%%%%%%%%%%%%%%%%%%%%%%%%%%%%%%%%%%%%%%
\subsection{Grasping of a Small Object}
\label{sec: small_offset_grasping_results}
%%%%%%%%%%%%%%%%%%%%%%%%%%%%%%%%%%%%%%%%%%%%%%%%%%%%%%%%%%%%%%%%%%%%%%%%%%%%%%%%
\begin{figure}[t]
	\centering
	\includegraphics[width=0.85\columnwidth]{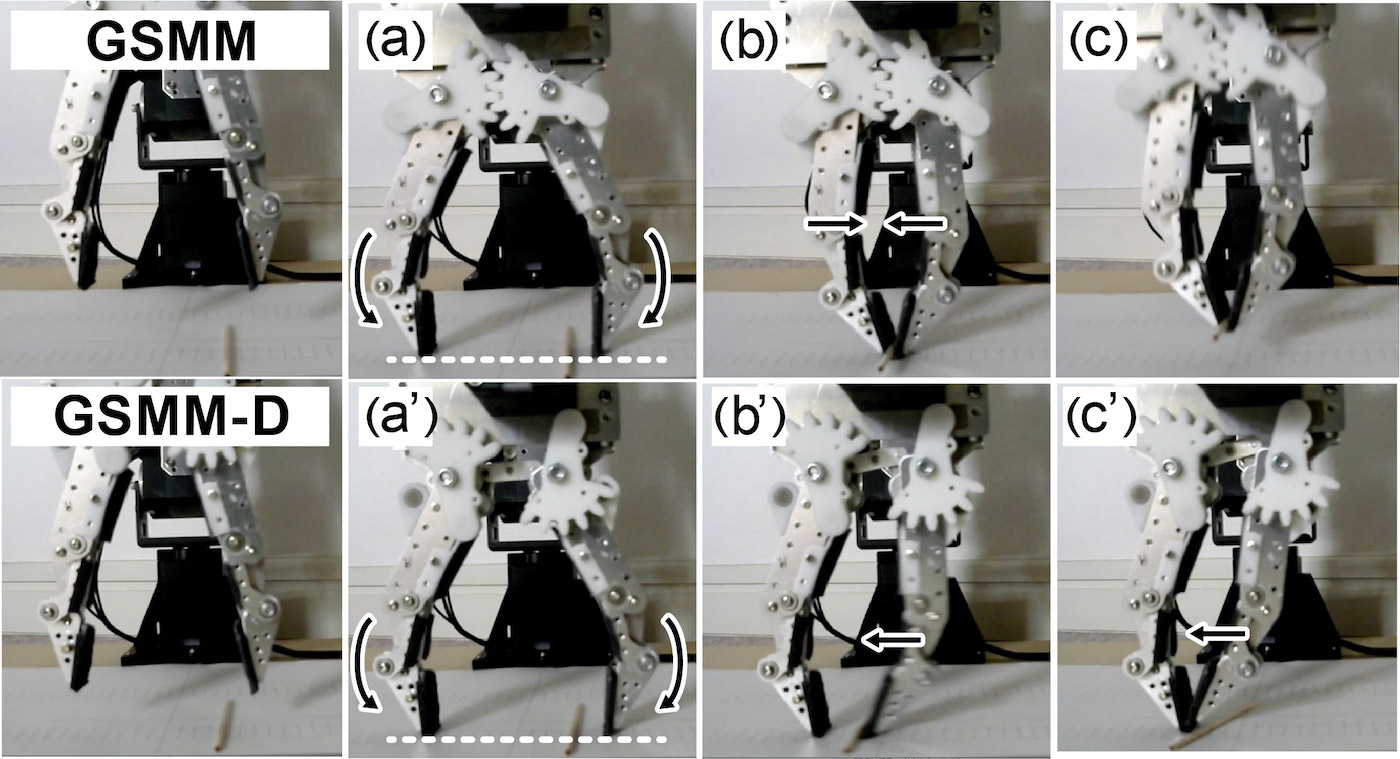}
	\caption{
	Results of grasping a toothpick with offset placement. 
    }
	\label{fig:result_offset_toothpick}
\end{figure}

The adaptive mechanism of the finger allows the fingertips to bend slightly when in contact with the surface the object is placed on (Fig.~\ref{fig:result_offset_toothpick}~(a)). This allows the fingertips to gently move the object towards the center of the gripper by sweeping it along the surface it was on. This mechanism also allows for imprecise positioning of the gripper on/above the workspace and is useful in the grasping of small objects.

With the GSMM mechanism engaged, the left and right fingers move in a coordinated manner towards the center, so the grasp can be made without any difference in heights of the fingertip when they are near-closed, so a pinch-grasp of potentially tiny objects is possible (Fig.~\ref{fig:result_offset_toothpick}~(b) \& (c)). With GSSM-D the left finger makes and remains in contact with the floor and does not move, so the grasp fails due to a difference in the position and height of the fingertips and the grasp fails~(Fig.~\ref{fig:result_offset_toothpick}~(b') \& (c')).

\begin{table}[t]
  \caption{Results of small object grasping}
  \label{table: offset_small_object}
  \centering
   \begingroup
   \scalefont{0.75}
  \begin{tabular}{cccc}
    \hline 

    Distance from center [mm]& F2 hand GSMM& F2 hand GSMM-D  & Parallel gripper \\
    \hline \hline
    0   & 10/10  & 10/10 & 10/10 \\
    5   & 10/10  & 10/10 & 10/10 \\
    10  & 10/10  & 10/10 & 10/10 \\
    15  & 10/10  & 4/10  & 10/10 \\
    20  & 10/10  & 0/10    & 10/10 \\
    25  & 10/10  & 0/10    & 10/10 \\
    \hline
  \end{tabular}
   \endgroup
\end{table}
 
Table~\ref{table: offset_small_object} shows the results for 10 grasp attempts of a toothpick placed at a distance of $0-25\mathrm{\,mm}$ (in increments of $5\mathrm{\,mm}$) from the center of the hand in the Y-axis direction in hand coordinate.
An attempt is considered successful when the object is lifted without being dropped.
The F2 hand with the proposed gear link mechanism (GSSM) had a 10/10 success rate as did the parallel gripper.
On the other hand, the F2 hand with the gear link mechanism disabled (GSSM-D) was unable to grasp a toothpick placed at a distance greater than $10\mathrm{\,mm}$.
This happens due to the left and right fingers not closing evenly and only one finger flexing significantly. This allows the toothpick to slide or roll under the finger and fail to be grasped.

%%%%%%%%%%%%%%%%%%%%%%%%%%%%%%%%%%%%%%%%%%%%%%%%%%%%%%%%%%%%%%%%%%%%%%%%%%%%%%%%
\subsection{Offset Grasping: Position changes}
\label{sec: weight_offset_grasping_results}
%%%%%%%%%%%%%%%%%%%%%%%%%%%%%%%%%%%%%%%%%%%%%%%%%%%%%%%%%%%%%%%%%%%%%%%%%%%%%%%%

%%%%%%%%%%%%%%%%%%%%%%%%%%%%%%%%%%%%%%%%%%%%%%%%%%%%%%%%%%%%%%%%%%%%
\subsubsection{Offset grasping of block-weights}
\label{sec: weight_offset_grasping_results}
%%%%%%%%%%%%%%%%%%%%%%%%%%%%%%%%%%%%%%%%%%%%%%%%%%%%%%%%%%%%%%%%%%%%

\begin{figure}[t]
	\centering
 	\includegraphics[width=0.99\columnwidth]{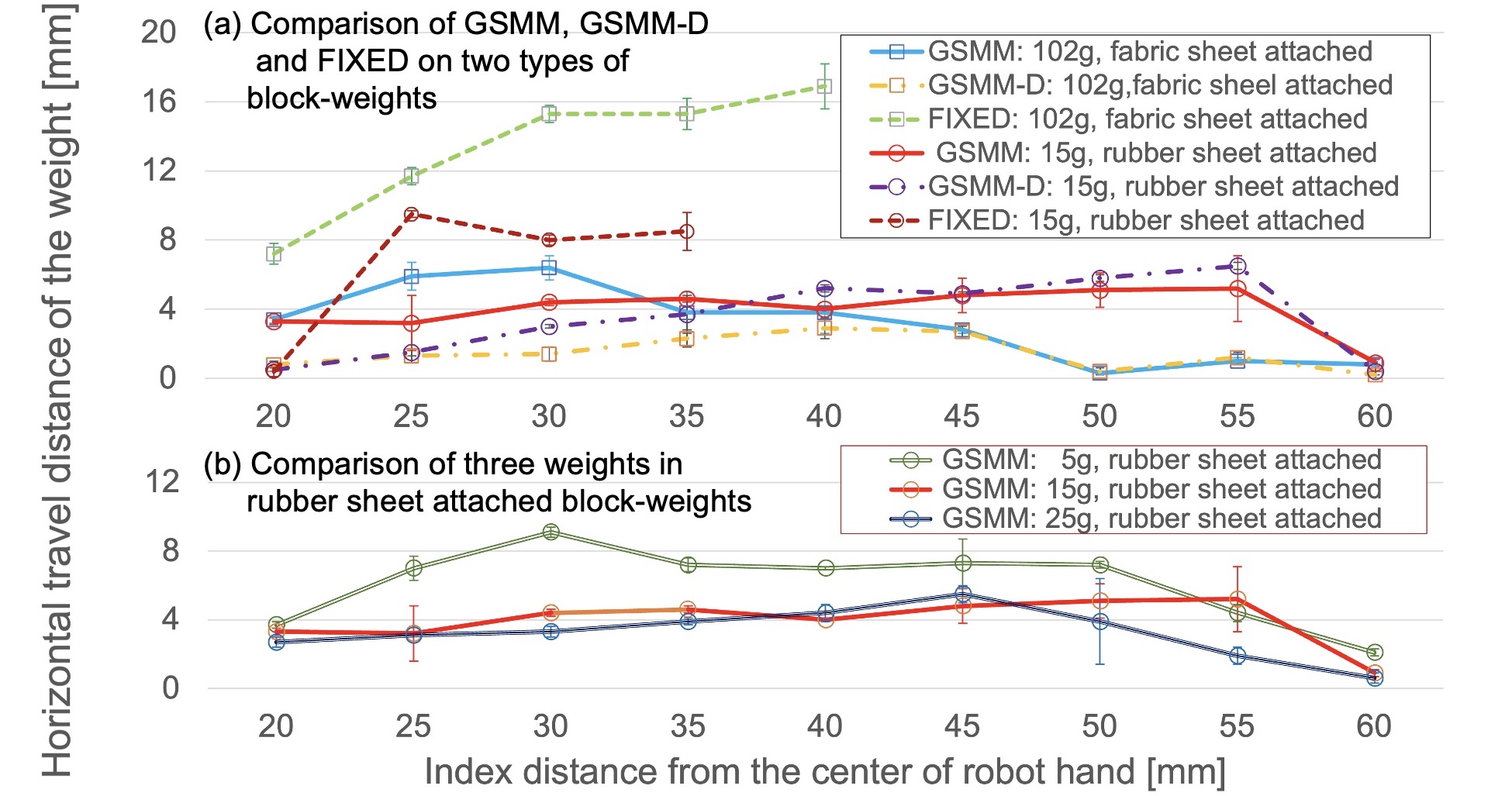}

	\caption{
	Results of offset grasping of the objects with three different grasping modes (GSMM, GSMM-D, and FIXED). 
    }
	\label{fig:result_weight_trans}
\end{figure}
As seen in Fig.~\ref{fig:result_weight_trans}~(a), we attempt to grasp a block-weight of $102\mathrm{\,g}$ with a fabric sheet to adjust its slipperiness and a $15\mathrm{\,g}$ case with a rubber sheet. Each experiment was repeated five times, and their displacement distances were recorded.
Under slippery conditions, lateral displacement occurred even for objects weighing as little as $102\mathrm{\,g}$. The largest behavior was observed at the position just before the magnet disengaged (and the GSMM switched to GSMM-D) ($35\mathrm{\,mm}$). Similarly, for FIXED, the weights tended to move significantly.
The difference between GSMM and GSMM-D in the $20-35\mathrm{\,mm}$ range is due to the force required to disengage the magnet. If the slightest lateral movement is also desired to be suppressed, GSMM-D can be chosen as the operating mode. This gives the user more options to expand the offset grasping performance of the hand if needed.
The GSMM-D, however, has issues as shown in Fig.~\ref{fig:badmotion}, so it is desirable to carefully examine the intended use before the decision is made.

With the rubber sheet attached, the lateral shift is slight, even though the weight is as light as $15\mathrm{\,g}$.  
In FIXED, as with the $102\mathrm{\,g}$ weight, the load is immediately transmitted to the object upon contact, and the object is greatly displaced laterally. The experiment was terminated when the weight exceeded $35\mathrm{\,mm}$ because the fingers would bend over before touching the object and could not grasp the weight.
The same experiment was conducted for $5\mathrm{\,g}$ and $25\mathrm{\,g}$ to confirm the effect of different weights (Fig.~\ref{fig:result_weight_trans}~(b). The $25\mathrm{\,g}$ showed the same tendency as the $15\mathrm{\,g}$ because they provided some frictional force, whereas the $5\mathrm{\,g}$ were too light and did not provide enough frictional force, and the weights tended to move significantly.
 These results indicate that the weight fluctuates during offset gripping under low friction conditions. This indicates that when actively using the offset grasping function, it is necessary to pay attention not only to the weight of the object but also to the state of friction.

%%%%%%%%%%%%%%%%%%%%%%%%%%%%%%%%%%%%%%%%%%%%%%%%%%%%%%%%%%%%%%%%%%%%
\subsubsection{Offset grasping of fragile objects}
\label{sec: strawberry_offset_grasping_results}
%%%%%%%%%%%%%%%%%%%%%%%%%%%%%%%%%%%%%%%%%%%%%%%%%%%%%%%%%%%%%%%%%%%%

\begin{figure}[t]
	\centering
	\includegraphics[width=0.87\columnwidth]{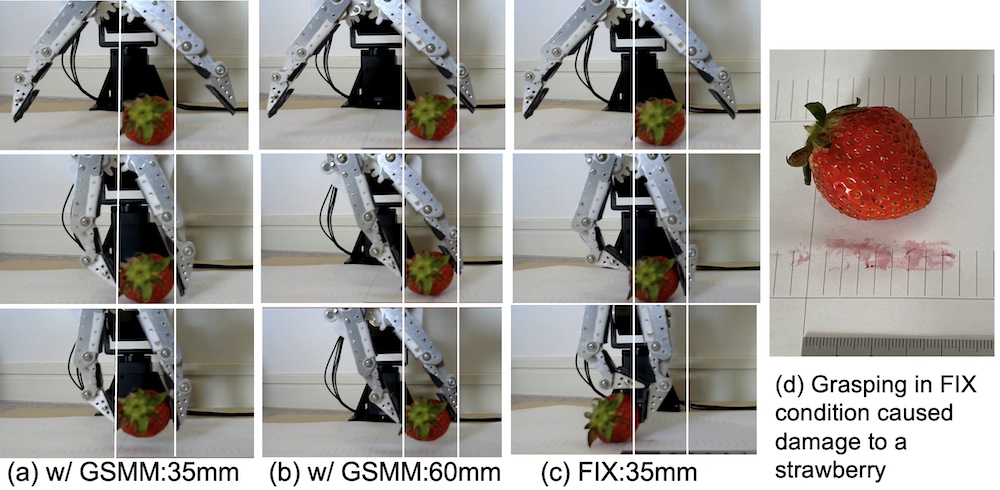}
	\caption{
	Results of offset grasping of strawberries. 
        }
	\label{fig:result_of_offset_strwwberry}
\end{figure}

To confirm whether a displacement of GSMM obtained in Sec.~\ref{sec: weight_offset_grasping_results} is acceptable for grasping fragile and easily damaged items like food, offset grasping of strawberries is performed.
For comparison, grasping is also performed with FIXED.
%The results are shown in Fig.~\ref{fig:result_of_offset_strwwberry}.
With GSMM, when the fingers contact the strawberry, the magnet tends to disengage and transition to an offset grasping motion before any load that would move or crush the strawberry (Fig.~\ref{fig:result_of_offset_strwwberry}). 
As a result, as shown in Fig.~\ref{fig:result_of_offset_strwwberry}~(a) \& (b), it could grasp the strawberry as it was without applying load until the rightmost edge moved $60\mathrm{\,mm}$ away from the gripper's center position, although it did sometimes change its posture slightly.
On the other hand, FIXED could not grasp the strawberry if it was displaced by more than $25\mathrm{\,mm}$. Fig.~\ref{fig:result_of_offset_strwwberry}~(c) shows the result at $35\mathrm{\,mm}$ where the grasp failed. The finger that contacted the strawberry generated a force that crushed the strawberry. This damage resulted in juice and pulp smearing the paper indicating that the strawberry was damaged (Fig.~\ref{fig:result_of_offset_strwwberry}~(d)).
These results confirm that the offset gripping function operates properly even with a small load, even when the magnetic gear linkage mechanism (GSMM) is initially engaged when appropriate thereby avoiding inflicting significant load or damage on the object.

\begin{figure}[t]
	\centering
	\includegraphics[width=0.83\columnwidth]{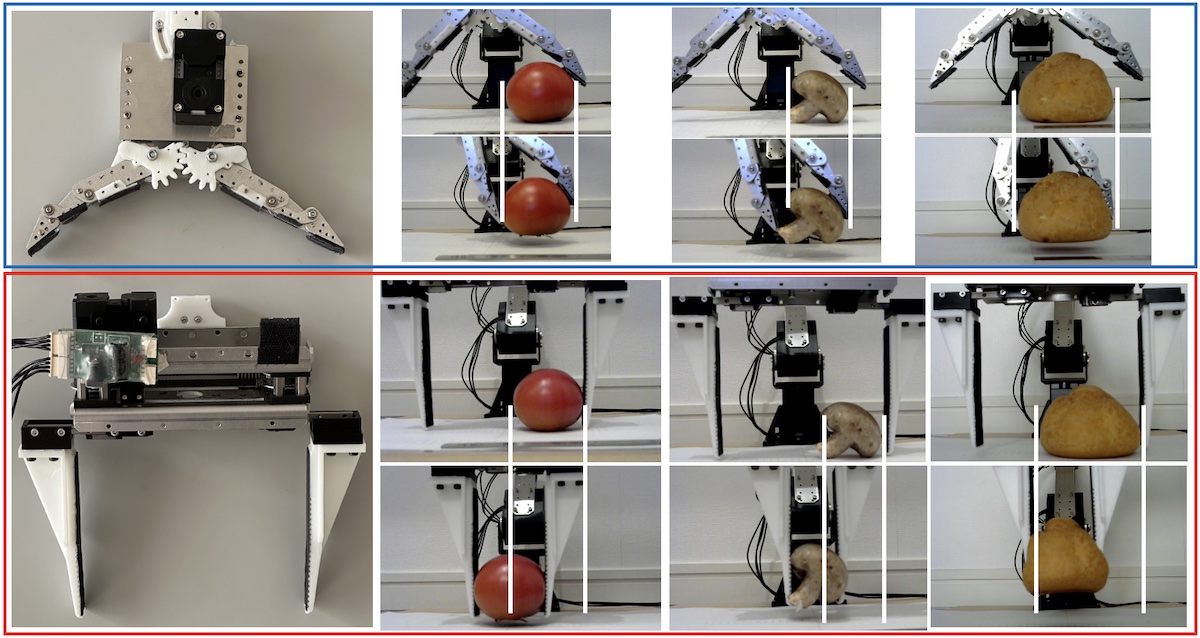}
	\caption{
    Example of fragile object grasping with offset grasping. Upper: Grasping by the F2 hand. Lower: Grasping by the parallel gripper.
    }
	\label{fig:fragile_ovject}
\end{figure}

Fig.~\ref{fig:fragile_ovject} shows the results of offset gripping experiments on tomatoes, shiitake mushrooms, and cream puffs using the F2 hand and parallel gripper.
With the parallel gripper, the object is first dragged along the surface and centered with the hand's axis (resulting in possible pose changes as seen with the mushroom) and picked up. With the F2 hand, the object is lifted with almost no variation in pose or position.
With grasping the cream puff, the parallel gripper lifts it in a significantly crushed state, while in the F2 hand, the degree of deformation is small. This is the effect of the compliance mechanism increasing the contact area thereby reducing pressure exerted.

%%%%%%%%%%%%%%%%%%%%%%%%%%%%%%%%%%%%%%%%%%%%%%%%%%%%%%%%%%%%%%%%%%%%%%%%%%%%%%%%
\subsection{Offset Peg-in-hole: Position and Posture Changes}
\label{sec: peg_in_hole_results}
%%%%%%%%%%%%%%%%%%%%%%%%%%%%%%%%%%%%%%%%%%%%%%%%%%%%%%%%%%%%%%%%%%%%%%%%%%%%%%%%

\begin{figure}[t]
	\centering
	\includegraphics[width=0.95\columnwidth]{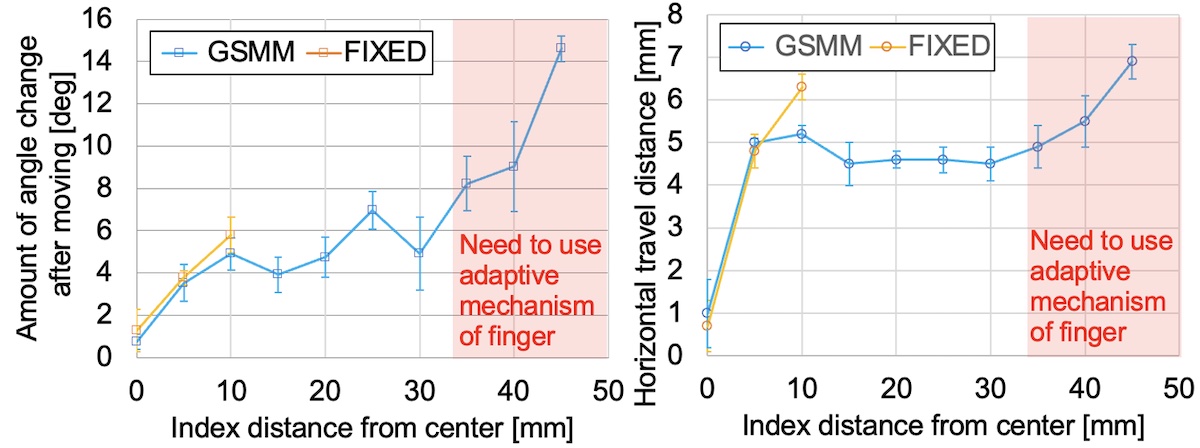}
	\caption{
	Results of horizontal displacement of the peg and rotation displacement when pulling up an offset peg. 
    }
	\label{fig:result_graph_of_peg-in-hole}
\end{figure}

\begin{figure}[t]
	\centering
	\includegraphics[width=0.85\columnwidth]{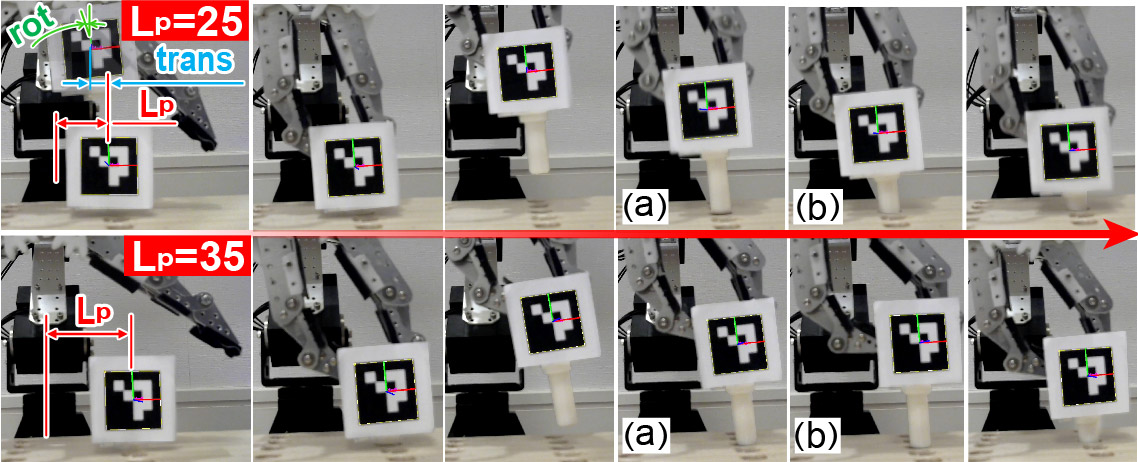}
	\caption{
	Peg-in-hole experiment to verify whether an inserted peg offset from the center of the robot hand by Lp could be grabbed, lifted, and reinserted. 
    }
	\label{fig:result_of_peg-in-hole}
\end{figure}

Fig.~\ref{fig:result_graph_of_peg-in-hole} shows the peg-in-hole experimental results. The offset grasping capability enabled the F2 hand to successfully pull out and insert back a peg that is axially offset by up to $45\mathrm{\,mm}$. There were no issues in pulling out and re-inserting the peg offset by $25\mathrm{\,mm}$, but with offsets above $30\mathrm{\,mm}$, the tilt angle of the peg became large. Typically, the peg can not be inserted at this angle because the edge hits the edge~(Fig.~\ref{fig:result_of_peg-in-hole} (a)), but the F2 hand's finger adaptive mechanism allows the peg to be jiggled side-to-side allowing it to be reinserted (Fig.~\ref{fig:result_of_peg-in-hole}~(b)). 
On the other hand, since the FIXED hand allows no independent finger movement and enforces coordinated movement, the tilt angle when the peg is removed is much larger, and at $15\mathrm{\,mm}$ or more, it tilts more than 45 degrees at the moment of removal. Therefore, the experiment was terminated at this point because it was not possible to reinsert the pin.

%%%%%%%%%%%%%%%%%%%%%%%%%%%%%%%%%%%%%%%%%%%%%%%%%%%%%%%%%%%%%%%%%%%%%%%%%%%%%%%%
\subsection{Autonomous Grasping}
\label{sec: autonomous_grasping_results}

%%%%%%%%%%%%%%%%%%%%%%%%%%%%%%%%%%%%%%%%%%%%%%%%%%%%%%%%%%%%%%%%%%%%%%%%%%%%%%%%

\begin{figure}[t]
	\centering
	\includegraphics[width=0.85\columnwidth]{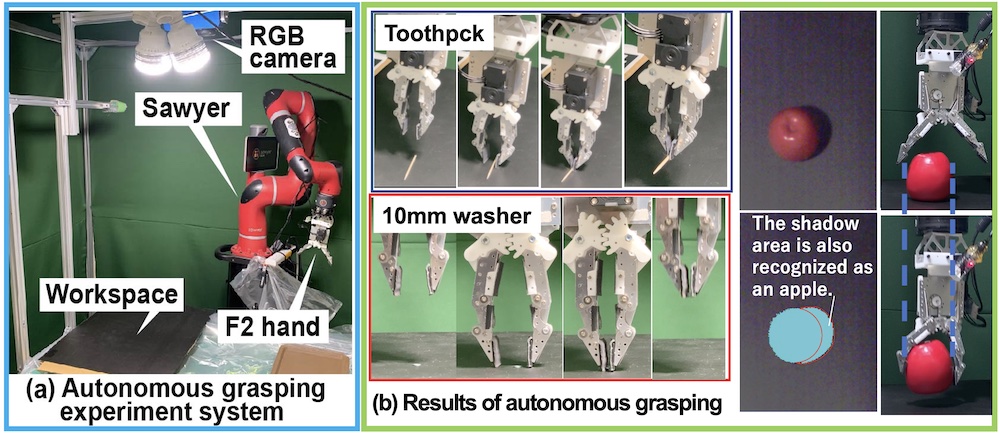}
	\caption{
        Results of autonomous grasping experiments.
    }
	\label{fig:shadow_detect}
\end{figure}

Using the experimental conditions described in Sec.~\ref{sec: autonomous_grasping_conditions}, we conduct experiments in which washers, toothpicks, and a plastic apple are automatically grasped by the robot.
Fig.~\ref{fig:shadow_detect}~(a) shows the experimental system and (b) shows the result of the grasp. 
The GSMM keeps both fingertips synchronized, and the adaptive mechanism of the fingers works when the fingertip touches the floor surface to execute the grasp. We verify that these mechanisms allow us to grasp difficult-to-grasp toothpicks and the $10\mathrm{\,mm}$ washer.
We also confirm that it is easy to grasp an apple, which along with the washer is a part of the YCB object dataset~\cite{ycb_data_2015}.

As the mobile manipulator moves, the environment and the robot's perception are constantly changing due to things like lighting, shadows, partial visibility, etc. Fig.~\ref{fig:shadow_detect}~(b) shows the recognition of an apple with a simple heuristic of background subtraction. While significant research deals with this problem, it is unlikely this can be fully compensated for. This intentionally simplistic choice of recognition demonstrates the case when the shadow is recognized as part of the grasp target. If we execute the grasp in this situation, the system aligns the hand to the centroid of the area (including the shadow). A normal parallel gripper spread wide is able to grasp the object while dragging it along the floor to the gripper's center during closure. On the other hand, F2's offset grasping function allows the gripper to automatically pick up the object in the same position without moving it. F2's functioning is therefore expected to improve the robustness of grasping even under severe recognition issues.

%%%%%%%%%%%%%%%%%%%%%%%%%%%%%%%%%%%%%%%%%%%%%%%%%%%%%%%%%%%%%%%%%%%%%%%%%%%%%%%%
\subsection{Teleoperation using the HSR}

%%%%%%%%%%%%%%%%%%%%%%%%%%%%%%%%%%%%%%%%%%%%%%%%%%%%%%%%%%%%%%%%%%%%%%%%%%%%%%%%
\subsubsection{Grasping wooden block}

\begin{figure}[t]
	\centering
	\includegraphics[width=0.90\columnwidth]{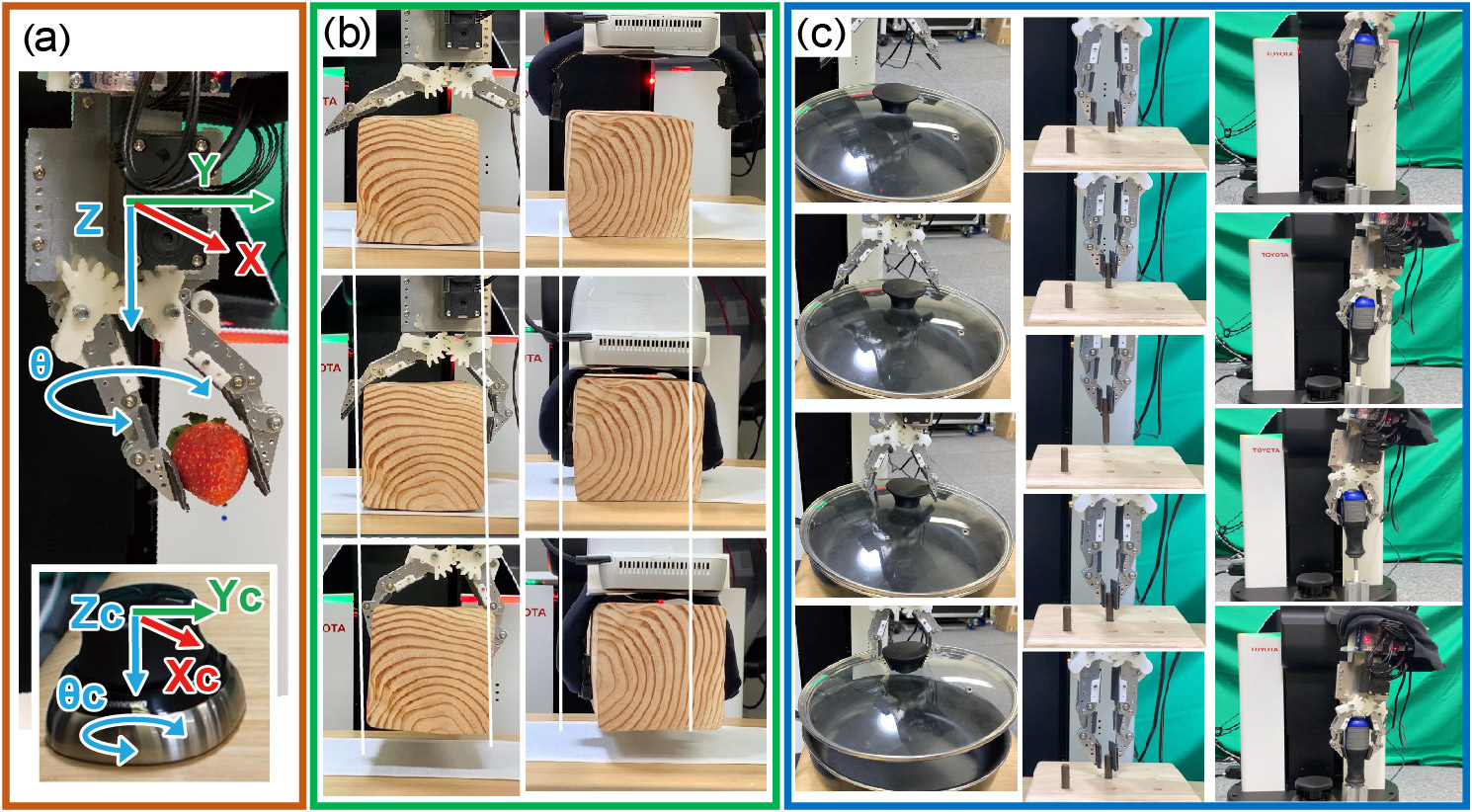}
	\caption{
    Confirmation of offset grasping function using HSR.
    }
	\label{fig:wood_block}
\end{figure}

When attempting to grasp a reasonably light object under teleoperation, even if the object is slightly misaligned from the arm, it can still succeed by closing its fingers and dragging the object along during closure. However, if the same is attempted with a heavy object resting on a surface, a significant amount of load is put on the gripper, the arm, the actuators, and the whole robot. This is especially true for mobile robots with a movable base. Typically, for such a grasp to succeed, either the arm, gripper or robot's base needs to be repositioned so the gripper and the object are aligned. 
If, however, the hand of the robot is able to perform offset grasping, it can absorb this misalignment and not require repositioning of the other parts of the robot. Fig.~\ref{fig:wood_block}~(b) shows snapshots of the grasping motion with the original HSR hand and F2 hand when grasping the block of wood (0.864 kg), which is one of the heaviest objects in the YCB objectset~\cite{ycb_data_2015}. The center of the wooden block is placed at 18mm offset from the hand center, which is the maximum displacement the HSR can grasp with its original hand. 
The maximum Y-axis force exerted on the HSR's wrist (measured by its force sensors) during the grasp was $4.57\mathrm{\,N}$ whereas the F2 was $0.47\mathrm{\,N}$ at the same amount of offset, and $0.99\mathrm{\,N}$ at the maximum displacement of $30\mathrm{\,mm}$, which was the maximum distance a successful offset-grasp the wood could be performed (Fig.~\ref{fig:wood_block}~(b)). This difference in force is attributed to the HSR grasping the wood by bringing it to the center of the palm, while F2 grasps the wood without moving it. 
The results show that without the offset grasping function, the load on the gripper increases when grasping a heavy object that is offset from the center of the robot hand. This increase in load can cause the actuator output to be insufficient depending on the amount of object offset, resulting in grasp failure.

%%%%%%%%%%%%%%%%%%%%%%%%%%%%%%%%%%%%%%%%%%%%%%%%%%%%%%%%%%%%%%%%%%%%%%%%%%%%%%%%
\subsubsection{Manipulation by teleoperation}

The offset grasping capability of the F2 hand also makes it easy to lift heavy, difficult-to-grasp objects such as frying pan lids, even with rough positioning (Fig.~\ref{fig:wood_block}~(c) left).
This offset grasping capability is also effective for common tasks such as peg-in-hole. When grasping a light peg (Fig.~\ref{fig:wood_block}~(c) middle), the offset grasping function allows for gentle grasping and insertion even when the alignment is imprecise. Fig.~\ref{fig:wood_block}~(c) right shows an example of inserting a screwdriver into a hole in an aluminum frame. The F2 hand is equipped with a compliance mechanism that allows offset grasping, so the posture of the grasped object can be changed as long as the object is not held with a strong force. This feature makes it possible to insert a peg even if it is caught by a peg by swinging the arm to the left or right.

%%%%%%%%%%%%%%%%%%%%%%%%%%%%%%%%%%%%%%%%%%%%%%%%%%%%%%%%%%%%%%%%%%%%%%%%%%%%%%%%
%%%%%%%%%%%%%%%%%%%%%%%%%%%%%%%%%%%%%%%%%%%%%%%%%%%%%%%%%%%%%%%%%%%%%%%%%%%%%%%%
\section{CONCLUSIONS}
In this paper, we explore the following two enhancements to expand the applicability of under-actuated robotic hands with an adaptive mechanism that operates with a single actuator.
1) Allowing the two fingertips to stay coordinated in position.
2) Allowing for independent adaptation of the fingers to the object's position in such a way as to allow for offset grasping while preserving the object's original pose after the grasp.
In order to achieve this, we propose a gear-type synchronization mechanism with a magnetic detachable structure in which each finger moves independently and adapts to the object to be grasped, and in which the movements of both fingers are synchronized until they contact the object.
We verify that our proposed robot hand can grasp small and thin objects such as a washer and toothpicks, and is capable of stable grasping of soft objects such as food. It can do this even if they are located far from the hand center and can do so while simultaneously avoiding damage by compliant grasping and offset grasping without dragging the object along into alignment. We have also verified the effectiveness of the synchronization mechanism for tasks that require precise positioning, such as peg-in-hole.

%%%%%%%%%%%%%%%%%%%%%%%%%%%%%%%%%%%%%%%%%%%%%%%%%%%%%%%%%%%%%%%%%%%%%%%%%%%%%%%%
%%%%%%%%%%%%%%%%%%%%%%%%%%%%%%%%%%%%%%%%%%%%%%%%%%%%%%%%%%%%%%%%%%%%%%%%%%%%%%%%
\section*{ACKNOWLEDGMENT}
The authors thank Shimpei Masuda, Dr. Tianyi Ko, and Dr. Koji Terada of, and formerly of Preferred Networks, Inc. and Koichi Ikeda, Hiroshi Bito, and Dr. Hideki Kajima from Toyota Motor Corporation for their assistance.
%%%%%%%%%%%%%%%%%%%%%%%%%%%%%%%%%%%%%%%%%%%%%%%%%%%%%%%%%%%%%%%%%%%%%%%%%%%%%%%%

\bibliographystyle{IEEEtran} %Line99 in IEEEtran to turn on et.al.
\bibliography{F2hand}

\end{document}